\documentclass[letterpaper, 10 pt, conference]{ieeeconf}  

\IEEEoverridecommandlockouts                              

\usepackage[english]{babel}
\usepackage[utf8]{inputenc}

\usepackage[colorinlistoftodos]{todonotes}

\newcommand{\eg}{\textit{e.g.}}
\newcommand{\ie}{\textit{i.e.}}
\newcommand{\etal}{\textit{et al.}}

\usepackage{amsmath, amsfonts, amssymb, amsthm}
\newtheorem{problem}{Problem}
\newtheorem{assumption}{Assumption}
\usepackage{graphicx}
\usepackage{float}
\usepackage{wasysym}  
\usepackage{booktabs} 
\usepackage{algorithm}
\usepackage{algpseudocode}
\usepackage{threeparttable}
\usepackage{url}

\title{\LARGE \bf
Can LLM Agents Solve Collaborative Tasks?\\A Study on Urgency-Aware Planning and Coordination

}

\author{João Vitor de Carvalho Silva \qquad Douglas G. Macharet
\thanks{This work was supported by CAPES/Brazil - Finance Code 001, CNPq/Brazil, and FAPEMIG/Brazil.}
\thanks{The authors are with the Computer Vision and Robotics Laboratory (VeRLab), Department of Computer Science, Universidade Federal de Minas Gerais, Brazil. E-mails: {\tt\small \{joaovitorcarvalho, doug\}@dcc.ufmg.br}.}%
}

\begin{document}

\maketitle
\thispagestyle{empty}
\pagestyle{empty}

\begin{abstract}
The ability to coordinate actions across multiple agents is critical for solving complex, real-world problems. Large Language Models (LLMs) have shown strong capabilities in communication, planning, and reasoning, raising the question of whether they can also support effective collaboration in multi-agent settings. In this work, we investigate the use of LLM agents to solve a structured victim rescue task that requires division of labor, prioritization, and cooperative planning. Agents operate in a fully known graph-based environment and must allocate resources to victims with varying needs and urgency levels. We systematically evaluate their performance using a suite of coordination-sensitive metrics, including task success rate, redundant actions, room conflicts, and urgency-weighted efficiency. This study offers new insights into the strengths and failure modes of LLMs in physically grounded multi-agent collaboration tasks, contributing to future benchmarks and architectural improvements.
\end{abstract}


\section{Introduction}
Coordinating actions across multiple agents is essential for solving real-world problems involving uncertainty, spatial constraints, and limited resources, 
especially in areas like
disaster response, robotic teams, and distributed systems.
Recent advances in artificial intelligence, particularly Large Language Models (LLMs), have introduced a new paradigm for building general-purpose agents capable of communication, planning, and reasoning~\cite{Guo2024LLM, Jin2025Comprehensive}. However, despite their growing capabilities, whether LLM-based agents can support robust and efficient coordination in dynamic, physically grounded environments remains an open research question~\cite{tran2025multiagentcollaborationmechanismssurvey, grötschla2025agentsnetcoordinationcollaborativereasoning}.



In this work, we investigate the coordination capabilities of LLM agents in the context of structured indoor victim rescue missions. In this scenario, each victim requires a specific type of aid (\eg, water, food, or medicine) and has a different level of urgency. Agents must collaborate to prioritize victims and deliver the appropriate resources efficiently. The environment is modeled as a graph, where nodes represent rooms or locations. Agents are given full knowledge of the environment, including the location and status of all victims, so the focus shifts from exploration to collaborative decision-making. Agents must decide where to go, which victims to assist, and how to divide responsibilities in order to maximize the number of victims helped while minimizing total steps.

Our goal is to assess whether LLM-based agents can exhibit emergent coordination behaviors, such as division of labor, urgency-aware planning, and cooperative task allocation. To this end, we systematically evaluate their performance using a suite of metrics, including: (i) number of victims rescued; (ii) redundant agent movements; (iii) room occupancy conflicts (instances where multiple agents enter the same location simultaneously); and (iv) urgency-weighted efficiency (steps taken relative to the priority of each type of victim). 
We additionally log detailed execution data to support analysis of coordination behavior, including miscommunications and hallucinated plans, phenomena also observed and analyzed in~\cite{park2023generativeagentsinteractivesimulacra}.




Whereas prior work on LLM-driven collaboration has focused mostly on symbolic reasoning or controlled games (\eg, bomb defusal or card games)~\cite{Li_2023}, our approach grounds the agents in a spatially and resource-constrained environment with real-world analogs. Inspired by recent studies on \emph{Theory of Mind} and emergent cooperation in LLMs, we extend the investigation to tasks involving physical navigation and urgency-based prioritization. 

To provide further insight, we introduce a rule-based heuristic baseline designed for task efficiency without language reasoning ~\cite{su2025dataefficientmultiagentspatialplanning,zheng2025llmmeetsskyheuristic}. Comparing LLM performance against this baseline highlights key strengths and limitations of LLM-based coordination, advancing research on emergent collaboration in grounded multi-agent systems~\cite{sun2025multiagentcoordinationdiverseapplications, grötschla2025agentsnetcoordinationcollaborativereasoning}.

%



\section{Related Work}




Recent work has explored the use of Large Language Models (LLMs) as zero-shot planners capable of translating natural language prompts into executable actions. For example, Da Silva~\etal~\cite{DaSilva2024Robotic} developed a ReAct-style agent that enables humanoid robots to reason about navigation and tool usage through LLM-driven reasoning chains. While effective for short tasks, their system suffered from decreased reliability over long-horizon plans due to context window limitations and hallucinated decisions. To address such issues, \cite{TaPA} introduced TaPA (Task Planning Agent), a task decomposition mechanism that aligns symbolic plans with visual perception, improving consistency in real-world execution. Collectively, these approaches highlight both the flexibility and current limitations of LLMs in robotic settings. 



Building on these capabilities, recent research has begun investigating the use of LLMs in multi-agent coordination problems. Park~\etal~\cite{park2023generativeagentsinteractivesimulacra} demonstrated that LLM-based agents in a sandbox environment could spontaneously exhibit cooperative behavior, including task allocation, scheduling, and role differentiation, even without predefined coordination protocols. Li~\etal~\cite{Li_2023} evaluated GPT-based agents in cooperative games, showing emergent Theory of Mind (ToM) behavior, with team performance comparable to reinforcement learning baselines. However, these studies also reveal persistent challenges, such as inconsistent dynamic belief states modeling, and hallucinated assumptions.


Several studies have proposed augmenting LLM agents with explicit ToM modules to improve coordination \cite{Oguntola2023Theory, Yuan2021Emergence}, while recent surveys \cite{Guo2024LLM, Jin2025Comprehensive, sun2025multiagentcoordinationdiverseapplications} have provided comprehensive overviews of LLM-based multi-agent systems. These works emphasize the potential of language-driven policies as a complement to traditional multi-agent reinforcement learning, especially in domains like disaster response, autonomous driving, and human-robot interaction. However, they also underscore the lack of systematic benchmarks in structured spatial environments where coordination demands are grounded in physical constraints.

Our study evaluates LLM agents in a graph-based rescue scenario. Agents must plan, act, and communicate to help victims with varying needs and urgency, while handling spatial constraints and avoiding redundancy. We use belief-state prompting to reduce hallucinations and include a deterministic heuristic for systematic comparison. This setup enables the analysis of emergent cooperation, task allocation, and spatial coordination, offering a novel testbed for high-level reasoning and multi-agent planning with LLMs.

\section{Problem formulation}

We model the rescue environment as an undirected graph $G = (V, E)$, where each node $v \in V$ represents a room, and each edge $(v_i, v_j) \in E$ denotes a bidirectional connection between adjacent rooms. Each room may contain at most \textbf{one} victim in need of assistance and maintains local connectivity information to support agent navigation and planning.

A set of victims $\mathcal{X} \subseteq V$ is distributed across the environment, each victim $x_v \in \mathcal{X}$ associated with:
\begin{itemize}
    \item A set of needs $N_v \subseteq \{\text{water}, \text{food}, \text{medicine}\}$;
    \item An urgency level $u_v \in \{\text{urgent}, \text{not\_urgent}\}$.
\end{itemize}

We define a team of agents $A = \{a_1, a_2, \dots, a_n\}$, where each agent has:
\begin{itemize}
    \item A position in the environment (\ie, current node in $V$);
    \item An inventory of resources, consisting of a limited number of units of water, food, and medicine used to assist victims; a resource can only be delivered if the corresponding quantity in the inventory is greater than or equal to one;
    \item The ability to make decisions autonomously based on its internal state and received communications.
\end{itemize}

\begin{assumption}
Agents have full knowledge of the environment, including the topology of $G$, the location of all victims, and each victim’s needs and urgency level. 
\end{assumption}

This setup removes the challenges of exploration and uncertainty, allowing us to isolate and study the agents’ capacities, coordination, planning, and resource allocation capabilities.

At each time step $t$, agents choose between two actions:
\begin{itemize}
    \item \emph{Move} to an adjacent room (\ie, a connected node in $E$);
    \item \emph{Deliver} a resource to a victim in the current room.
\end{itemize}

To facilitate coordination, agents are required to communicate at every decision step. After taking an action, each agent generates a message, summarizing its recent activity, intentions, or observations, and broadcasts it to a shared message channel accessible to all agents. 


\begin{problem}
The problem is to evaluate the ability of LLM-driven agents to achieve effective coordination in a fully observable, resource-constrained multi-agent rescue task. Each agent objective is to maximize the number of victims who are fully assisted, \ie, all their required resources are delivered by one or more agents. Formally, the team objective is to maximize the following cumulative reward function:
%
%
\begin{equation}
f(A,\mathcal{X}) = \sum_{v \in \mathcal{X}} s(x_v),
\end{equation}
where $s(x_v) = 1$ if all needs $N_v$ of victim $x_v$ are satisfied, and $s(x_v) = 0$ otherwise.
\end{problem}



\section{Methodology}


This section details the architectural design and operational logic of the agents used in the simulation framework. Two agent types are implemented: (i)~LLM-driven agents based on modular reasoning graphs, and (ii)~deterministic heuristic agents that follow a fixed policy and serve as a performance baseline.

\subsection{LLM-guided Planning and Action}

\begin{figure}[htpb]
    \centering
    \includegraphics[width=\linewidth]{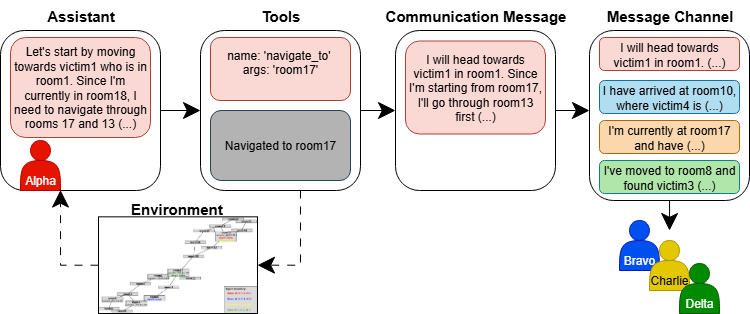}
    \caption{ReAct-based decision pipeline for LLM-driven agents.}
    \label{fig:pipelineRescueAgent}
\end{figure}

An overview of the reasoning pipeline is shown in Fig.~\ref{fig:pipelineRescueAgent}.

\textbf{LLM Agent Architecture: } The LLM-driven agents are implemented using Ollama as the provider of the models, 
and LangGraph, with a modular state graph architecture composed of three primary nodes responsible for a specific role in the agent's reasoning and acting process:

\begin{itemize}
    \item {\texttt{Assistant}}: Responsible for reasoning over the current internal knowledge and external observations, and determining the next action the agent should perform. 

    \item {\texttt{Tools}}: Executes the action selected by the assistant node. Available tools include navigation commands and victim assistance actions.
    
    \item {\texttt{Communication Message}}: Generates communication messages that are broadcasted to other agents. 
\end{itemize}

\textbf{Decision Cycle:}
At each decision step, an agent observes the environment, the communication channel, and its internal state, selects an appropriate action from its available tools, and broadcasts a communication message. The available tools include \texttt{navigate\_to(room)} for movement, \texttt{give\_water()}, \texttt{give\_food()}, and \texttt{give\_medicine()} for delivering resources, and \texttt{end\_mission()} to conclude operations.

\textbf{Communication Mechanism:}
Agents communicate through messages prefixed with \texttt{communicate:} posted to a shared message channel visible to all active agents. These messages summarize recent actions, intended plans, or observations about the environment. Each message has an expiration time of one step to ensure timely updates and prevent stale information from influencing decisions. This design promotes dynamic coordination and implicit task allocation among agents.

An example of such message is:
\begin{quote}
\texttt{communicate: I've now helped victim1 in room1. Since Alpha is heading toward victim5 at room13, and Charlie is going to victim2 at room4, I'll move to help victim3 at room7.}
\end{quote}

\subsection{Deterministic Heuristic Policy}
\label{sec:heuristic_agent}

A heuristic policy (Alg.~\ref{alg:heuristic_policy}) is implemented as a deterministic baseline to assess the performance of LLM agents. This policy simulates a multi-agent setup by being \textit{instantiable}, allowing each agent configuration to be mapped directly to a policy instance. The agent behavior follows a fixed set of rules that prioritize efficiency in resource delivery without any linguistic reasoning or adaptive communication.








\begin{algorithm}[h]
\small
\caption{Heuristic Agent Policy (per step)}
\label{alg:heuristic_policy}
\begin{algorithmic}[1]
\Require Agent state 
, victims $V$, agents $A$

\If{inactive or (no target and unable to help anyone)} \Return \EndIf

\If{no target}
    \State $C \gets$ victims needing assistance that agent can help
    \State Remove $v \in C$ if a closer agent can fully assist
    \If{$C = \emptyset$} \State Set inactive; \Return \EndIf
    \State Select target $v \in C$ with max help score and min distance
\EndIf

\If{not in target room}
    \State Move one step toward target
    \Return
\EndIf

\State Deliver one matching item
\If{target fully assisted or agent cannot assist further} \State Clear target \EndIf
\end{algorithmic}
\end{algorithm}

The agent iteratively selects a target victim, navigates towards them, and attempts to assist based on available resources. Victim selection is based on the number of needs the agent can fulfill and the distance to the victim's room. Victims that are already fully serviceable by another agent are avoided to reduce redundancy. Navigation is performed using Dijkstra's algorithm to ensure optimal movement between rooms. The process continues until all assistable victims are helped or the agent’s resources are exhausted.


The heuristic policy nature allows for the establishment of a strong deterministic baseline, highlighting the added value and challenges of language-informed decision-making in more complex agent architectures.

\section{Experimental Setup}


\subsection{Evaluation Protocol}


We evaluate the coordination capabilities of LLM-based agents in a structured, graph-based rescue environment. Each test instance is defined by a combination of a map layout, victim distribution, and agent configuration. We use 8 distinct maps, along with 8 victim configurations and 12 agent setups.


For each configuration, we execute the task using:
\begin{itemize}
    \item Each of the 8 LLM models with temperature $T=0.0$;
    \item The same models with temperature $T=0.5$;
    \item A deterministic rule-based heuristic policy (Sec.~\ref{sec:heuristic_agent}). 
\end{itemize}

Temperature controls the randomness of the model's decisions, with lower values yielding more deterministic behavior and higher values encouraging exploration.


Each setup is run once per model-temperature pair, totaling three runs per environment configuration, and execution logs are saved.

\subsection{Language Models}

We evaluate eight LLMs with varying parameter sizes, grouped as follows: Cogito, Qwen3, and Qwen2.5 with 14 billion parameters; Mistral-Small with 24 billion parameters; and larger variants including Cogito, Qwen3, Qwen2.5, and Qwen2.5-Coder with 32 billion parameters.





\subsection{Evaluation Metrics}

To evaluate agent performance and coordination, we gather quantitative metrics from each run, capturing aspects such as redundancy, efficiency, urgency responsiveness, and task allocation:

\begin{itemize}
    \item \textbf{final\_victims\_amount}: Number of victims not fully assisted by the end of the execution.
    \item \textbf{num\_steps}: Total number of steps taken by all agents (a step is a complete cycle of all agents turns).
    \item \textbf{total\_redundant\_agent\_moves}: Number of times an agent returned to a room it had previously visited.
    \item \textbf{steps\_2\_or\_more\_agents\_same\_room}: Total number of steps with two or more agents in the same room.
    \item \textbf{occurrences\_2\_or\_more\_agents\_same\_room}: Total number of distinct occurrences where multiple agents shared the same room.
    \item \textbf{average\_steps\_attend\_urgent\_victims}: Average number of steps taken to reach and assist urgent victims.
    \item \textbf{average\_steps\_attend\_not\_urgent\_victims}: Average number of steps taken to assist non-urgent victims.
\end{itemize}


\subsection{Baselines and Comparisons}

We compare the LLM-guided agents to a deterministic heuristic policy (Sec.~\ref{sec:heuristic_agent}) to isolate the value of language-driven reasoning. 
%

\subsection{Experimental Objectives}

Experiments aim to answer the following key questions:
%

\begin{enumerate}
\item \textbf{Collaborative Performance:} Can multiple LLM-based agents coordinate actions to reach shared goals?
\item \textbf{Spatial Reasoning:} Can agents understand the graph structure and plan efficient routes?
\item \textbf{Coordination Quality:} How well do agents divide tasks in multi-agent settings?
\item \textbf{Effect of Temperature:} How does decoding temperature influence coordination and task success?
\item \textbf{Impact of Prompting:} How do prompt variations affect prioritization, especially in scenarios with urgency?
\item \textbf{Comparison to Heuristics:} Can LLM-based agents match or surpass heuristic policies in challenging rescue scenarios?
\end{enumerate}

\section{Results and discussion}

    In this section, we analyze the performance of different models in rescue scenarios. It is important to note that direct quantitative comparison between successful and unsuccessful missions is not straightforward. Failures can occur due to distinct reasons, such as exceeding the maximum number of steps (60), early mission termination by agents, or detection of infinite loops. Thus, for failed missions, a qualitative analysis is more appropriate to understand agent behavior and limitations.


    We adopt shortened labels for all models throughout the text and figures for conciseness. The abbreviations are as follows: c14 = Cogito:14b, q14 = Qwen2.5:14b, m24 = Mistral-small:24b, c32 = Cogito:32b, q32 = Qwen2.5:32b, q32c = Qwen2.5-coder:32b, q3-14 = Qwen3:14b, and q3-32 = Qwen3:32b.

    \subsection{Collaborative Performance} 
\begin{figure}[H]
        \centering
        \includegraphics[width=\linewidth]{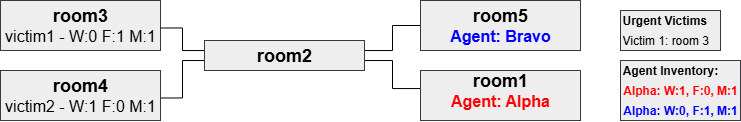}
        \caption{Instance setup \texttt{[map\_7, agents\_1, victims]} to evaluate a two agent cooperative rescue scenario.}
        \label{fig:map_7-agents_1-victims}
    \end{figure}

To evaluate if two LLM-based agents can coordinate movement to achieve a rescue goal, the map setup shown in Figure~\ref{fig:map_7-agents_1-victims} can be used. In that case, each agent possess the exact set of aids that one of the victims need. The victim in room 3 has urgent needs, while the victim in room 4 does not: this will allow us to see if agent Alpha, who takes action first, will coordinate with Bravo (by letting him take care of the urgent victim), or will chose the victim without reasoning about Bravo's resources.

    Half of the models couldn't save all the victims. The models that serviced all the victims did it within 1.20x to 2.5x steps compared to the heuristic baseline.

    \begin{itemize}

        \item Most models (\textbf{Cogito:32b}, \textbf{Qwen2.5-coder:32b}, and \textbf{Cogito:14b} at temperature 0.0), failed to optimize their strategy, choosing victims whose needs were not aligned with their inventories. This suboptimal decision prevented the agents from completing the mission successfully:
        \begin{itemize}
            \item \textbf{Cogito:14b (temperature 0.0)}: The agents failed to coordinate. Bravo ended the mission prematurely, while Alpha, after some effort, managed to use his resources.
            \item \textbf{Cogito:32b (temperature 0.0)}: The mission was ended prematurely without fully utilizing the agents' potential.
            \item \textbf{Qwen2.5-coder:32b (temperature 0.0)}: The agents attempted to adjust their initial plan, but Bravo hallucinated having helped a victim and failed to assist someone who could still be saved, leading to mission termination.
            \item \textbf{Cogito:32b (temperature 0.5)}: One agent successfully reached and helped another victim, but the other ended the mission prematurely.
            \item \textbf{Qwen2.5-coder:32b (temperature 0.5)}: After the initial suboptimal choice, the agents got stuck in a loop of thought and could no longer act.
        \end{itemize}

        \item \textbf{Qwen2.5:14b (temperature 0.0)}: The agents struggled with navigation, entering wrong rooms or staying idle, and ultimately ended the mission with one aid still missing.
        
        \item \textbf{Cogito:14b (temperature 0.5)}: The agents moved together, realized they could split up after helping one victim, but then chose the wrong path to the next and terminated the mission.
        
        \item \textbf{Qwen3:32b (temperature 0.5)}: The agents also moved together initially and then got stuck in a loop of thought, becoming unable to act.
    
    \end{itemize}


Table~\ref{tab:model_performance_test_2} reports task success or failure for each model.


\begin{table}[H]
\centering
\caption{Success (\CIRCLE) or failure (\Circle) of models in the rescue task.}
\begin{tabular}{lcccccccc}
\toprule
\textbf{Temp.} & \textbf{c14} & \textbf{q14} & \textbf{m24} & \textbf{c32} & \textbf{q32} & \textbf{q32c} & \textbf{q3-14} & \textbf{q3-32} \\
\midrule
\textbf{0.0} & \Circle & \Circle & \CIRCLE & \Circle & \CIRCLE & \Circle & \CIRCLE & \CIRCLE \\
\textbf{0.5} & \Circle & \CIRCLE & \CIRCLE & \Circle & \CIRCLE & \Circle & \CIRCLE & \Circle \\
\bottomrule
\end{tabular}
\label{tab:model_performance_test_2}
\end{table}

Although some models succeeded, success did not always imply optimal planning. For example, the model \textbf{qwen2.5:32b (temperature 0.0)} showed 7 instances of two or more agents occupying the same room, compared to only 1 for the heuristic baseline. This reveals a coordination issue: agents failed to divide tasks efficiently, with one often following another, causing unnecessary overlap. While all victims were ultimately helped, this behavior reduced mission efficiency, increasing the total steps required.

    \subsection{Spatial Reasoning} 
\begin{figure}[H]
        \centering
        \includegraphics[width=\linewidth]{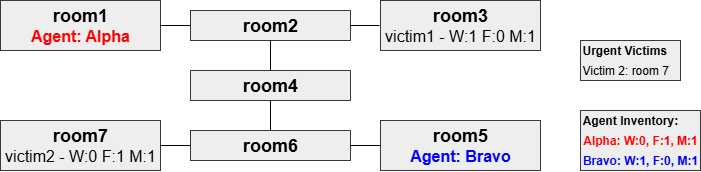}
        \caption{Instance setup \texttt{[map\_8, agents, victims]} to evaluate a two agent cooperative rescue scenario.}
        \label{fig:map_8-agents-victims}
    \end{figure}

To assess whether LLM-based agents understand graph structure and optimize routes, we examine the map in figure \ref{fig:map_8-agents-victims}. In this setup, each agent carries the exact aids required by one specific victim. However, reaching that victim requires crossing the entire map while ignoring a closer, but mismatched, victim. This will allow us to see if the agents can make long-distances plans, of if they blindly chose near victims.

    In this test setup, most instances were completed using fewer than twice the number of steps compared to the heuristic baseline, demonstrating a capacity for coordination in such scenarios despite occasional difficulties. Although the \textbf{Qwen2.5:14b (temperature 0.0)} successfully completed the mission, one of the agents had huge difficulty reasoning about the environment. Its logs indicate that it was unable to figure out how to reach a room containing a victim, resulting in a test run that required 10 times more steps than the heuristic policy.
    
    Despite the overall results, some models exhibited failures during the test:

    \begin{itemize}
    \item \textbf{Qwen3:14b (temperature 0.0)}: The agents ended the execution too early.
    \item \textbf{Mistral-small:24b (temperature 0.0)}: The agents had trouble to understand the environment, and the execution detected a malicious loop.
    \item \textbf{Qwen2.5:32b (temperature 0.0)}: After making not optimal choices about the victim saving, they ended the execution missing one aid.
    \item \textbf{Cogito:14b (temperature 0.5)}: The agents did well, but missed one aid and ended the execution too early.
    \item \textbf{Qwen3:14b (temperature 0.5)}: One of the agents got lost and couldn't deliver the last aid.
    \item \textbf{Qwen2.5:14b (temperature 0.5)}: One of the agents was lost the entire run, and couldn't use its resources.
    \item \textbf{Qwen3:32b (temperature 0.5)}: The agents coudn't deliver all the resources due to lack of coordination.
    \end{itemize}

    Table~\ref{tab:model_performance_test_3} summarizes the success of failure of the models in the task.


    \begin{table}[H]
    \centering
    \caption{Success (\CIRCLE) or failure (\Circle) of models in the rescue task.}
    \begin{tabular}{lcccccccc}
    \toprule
    \textbf{Temp.} & \textbf{c14} & \textbf{q14} & \textbf{m24} & \textbf{c32} & \textbf{q32} & \textbf{q32c} & \textbf{q3-14} & \textbf{q3-32} \\
    \midrule
    \textbf{0.0} & \CIRCLE & \CIRCLE & \Circle & \CIRCLE & \Circle & \CIRCLE & \Circle & \CIRCLE \\
    \textbf{0.5} & \Circle & \Circle & \CIRCLE & \CIRCLE & \CIRCLE & \CIRCLE & \Circle & \Circle \\
    \bottomrule
    \end{tabular}
    \label{tab:model_performance_test_3}
    \end{table}

    \subsection{Coordination Quality} To evaluate the coordination capabilities of the agents, we selected the scenario shown in figure 6. This configuration was deemed suitable for assessing task division and collaborative planning. We analyzed logs from both successful and unsuccessful executions and categorized coordination into three levels: \textbf{0 (no coordination)}, \textbf{1 (partial coordination)}, and \textbf{2 (high coordination)}.

     \begin{figure}[H]
        \centering
        \includegraphics[width=\linewidth]{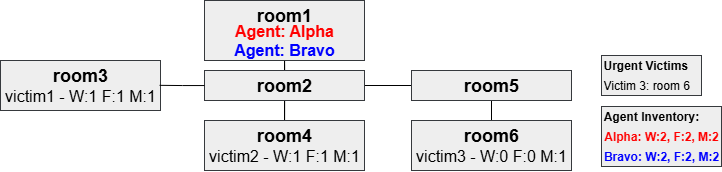}
        \caption{Instance setup \texttt{[map\_2, agents, victims]} to evaluate coordination.}
        \label{fig:map_2-agents-victims}
    \end{figure}

    In the best-performing case (\textbf{Cogito:32b 0.5 temperature}), agents displayed level~2 coordination. Agent Alpha assigned the remaining victims to Bravo, who correctly followed the plan. Communication was frequent and accurate, with Alpha adjusting actions based on Bravo's updates.
    
    Another level~2 example (\textbf{Cogito:32b 0.0 temperature}) showed similar delegation: Alpha handled the urgent victim and correctly assumed Bravo would handle the others. Minimal redundant actions occurred.
    
    Level~1 coordination was observed in \textbf{Qwen2.5:32b (temperature 0.5)}, where agents mostly divided the victims well, but briefly overlapped in the final room due to lack of awareness of each other's progress.

    Conversely, \textbf{Cogito:14b (temperature 0.0)} demonstrated level~0 coordination despite completing the task quickly. Each agent acted independently, with minimal communication and unintentional alignment.
    
    In contrast \textbf{Qwen3:32b (temperature 0.5)} and \textbf{Mistral-small:24b (temperature 0.0)} exhibited poor coordination. Agents often duplicated efforts or misunderstood task completion status. In both cases, the lack of role assignment and reliable communication led to unnecessary delays or failure to assist all victims, representing level~0 coordination.
    
    These findings suggest that while some LLM-based agents can plan and collaborate effectively, others fail to share intentions or adapt based on teammate actions, highlighting coordination as a key challenge in multi-agent scenarios.

    \subsection{Effect of Temperature} 
    To assess the impact of decoding temperature on coordination, we compare success rates at $T=0.0$ (deterministic) and $T=0.5$ (moderate randomness). The difference is minimal: agents rescued 188 victims at $T=0.0$ (67.1\%) and 191 at $T=0.5$ (68.2\%). These results suggest that a modest increase in temperature does not harm performance and may slightly improve coordination, though further experiments would be needed for stronger conclusions.

    

    \subsection{Impact of Prompting}

\begin{figure}[h]
        \centering
        \includegraphics[width=\linewidth]{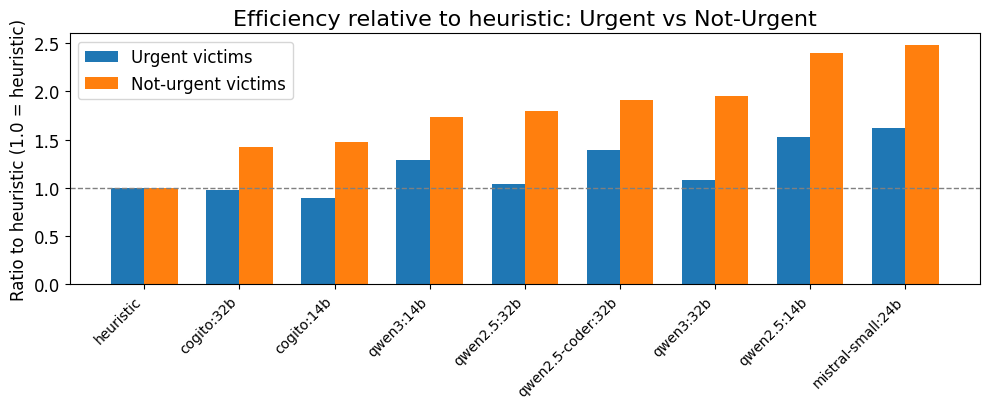}
        \vspace{-7mm}
        \caption{
        Efficiency of LLM models relative to the heuristic baseline when assisting both urgent and non-urgent victims. Values below 1.0 indicate better efficiency (fewer steps) than the heuristic. All models show lower ratios for urgent victims compared to non-urgent ones, demonstrating that LLMs more effectively prioritized urgent cases compared to the baseline.
        }
        \label{fig:efficiency_relative_to_heuristic_urgent_vs_not_urgent}
        \vspace{-2mm}
    \end{figure}

    To assess the impact of prompting on decision-making, we measured agents’ efficiency in assisting victims, leveraging the fact that urgency information is explicitly encoded in the prompt. We computed urgency-weighted average steps per victim type, normalizing by frequency across configurations to prevent bias, and compared results against a heuristic baseline that uses urgency only as a tie-breaker. Direct comparisons between urgent and non-urgent cases were avoided, as victim locations may vary significantly.
    As shown in Figure~\ref{fig:efficiency_relative_to_heuristic_urgent_vs_not_urgent}, all models consistently outperformed the heuristic in assisting urgent victims, suggesting they correctly interpreted urgency cues and prioritized accordingly. These findings indicate that LLMs can effectively exploit task-relevant textual information to adapt their strategies, enhancing multi-agent coordination through improved prioritization.

    \subsection{Comparison to Heuristics}



\begin{figure}[h]
    \centering
    \includegraphics[width=\linewidth]{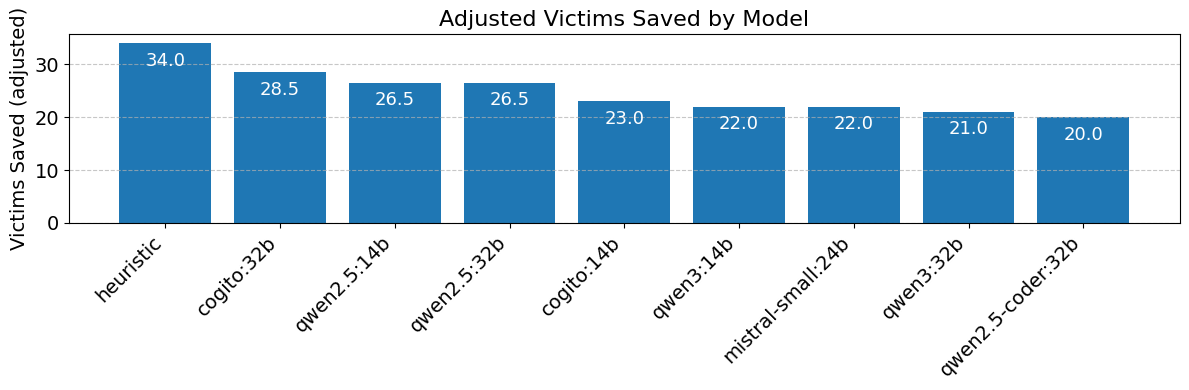}
    \vspace{-6mm}
    \caption{
    Total victims rescued per model, with LLM scores averaged over temperatures 0.0 and 0.5.
    }
    \label{fig:comparison_to_heuristics}
\end{figure}

We compared LLM-based agents to a deterministic heuristic by measuring the total number of victims rescued. Since each LLM was evaluated using two decoding temperatures (0.0 and 0.5), their scores were averaged to ensure fair comparison with the heuristic, which was run once per scenario.
Figure~\ref{fig:comparison_to_heuristics} shows the normalized results. The heuristic outperformed all LLM models, rescuing 34 victims in total. The best LLM, Cogito:32b, saved 28.5 victims, while Qwen2.5-coder:32b had the lowest performance with 20 victims rescued.

\vspace{-2mm}

Although no LLM surpassed the heuristic, top models approached its performance in terms of victims rescued, showing they can achieve near-baseline outcomes when considering the total number of victims saved.

\section{Conclusion and Future Work}

This work investigated whether LLM-based agents can effectively coordinate in structured multi-agent rescue tasks requiring urgency-aware planning, division of labor, and communication. Our evaluation across diverse models showed that, while some LLM agents exhibited promising coordination and prioritization, they still underperformed compared to a deterministic heuristic baseline in efficiency and reliability.

Key challenges identified include hallucinated plans, premature termination, and redundant actions, often due to limited awareness of teammates’ intentions and insufficient spatial reasoning. Nonetheless, top-performing LLMs demonstrated emergent coordination and consistently prioritized urgent cases better than the heuristic-driven agents.

Future work focus on improving belief-state tracking and shared world models, potentially via explicit memory to reduce hallucinations and redundancy. Evaluations in noisy environments would further assess robustness. We also advocate for standardized benchmarks in grounded multi-agent collaboration to facilitate consistent comparisons across emerging agent architectures.




\bibliographystyle{IEEEtran}
\bibliography{bibliography}

\end{document}